\documentclass{article} 
\usepackage{iclr_set,times}

\usepackage{amsmath,amsfonts,bm}

\def\eqref#1{equation~\ref{#1}}

\def\1{\bm{1}}

\DeclareMathAlphabet{\mathsfit}{\encodingdefault}{\sfdefault}{m}{sl}
\SetMathAlphabet{\mathsfit}{bold}{\encodingdefault}{\sfdefault}{bx}{n}

\usepackage{hyperref}
\usepackage{url}
\usepackage{graphicx}
\usepackage{booktabs}
\usepackage{multirow}
\usepackage{comment}
\usepackage{array} 
\newcolumntype{P}[1]{>{\raggedright\arraybackslash}p{#1}} 

\title{The Effect of Model Size on LLM Post-hoc \text{Explainability} via LIME}

\iclrfinalcopy

\author{%
        Henning Heyen\thanks{Contact:  \texttt{henning.heyen.22@ucl.ac.uk}}, Amy Widdicombe, Noah Y. Siegel, Mar\'{i}a P\'{e}rez-Ortiz, Philip Treleaven\\  
        \hspace{4.8cm}University College London\\
}

\begin{document}

\maketitle
\begin{abstract}
Large language models (LLMs) are becoming bigger to boost performance. However, little is known about how explainability is affected by this trend. This work explores LIME explanations for DeBERTaV3 models of four different sizes on natural language inference (NLI) and zero-shot classification (ZSC) tasks. We evaluate the explanations based on their \textbf{faithfulness} to the models'  internal decision processes and their \textbf{plausibility}, i.e. their agreement with human explanations. The key finding is that increased model size does not correlate with plausibility despite improved model performance, suggesting a misalignment between the LIME explanations and the models' internal processes as model size increases. Our results further suggest limitations regarding faithfulness metrics in NLI contexts. 

\end{abstract}

\section{Introduction}

Research has shown that performance in language models depends strongly on scale and less on model shape \citep{kaplan}, where scale refers to the number of parameters, the training dataset size, and the amount of compute for training. For instance, \mbox{OpenAI's} series of Generative Pre-Trained Transformers (GPT) has grown from 1.5 billion parameters for GPT-2 to 175 billion parameters for GPT-3 which helped improve across various NLP tasks \citep{brownGPT3}. This trend seems likely to continue. 

As LLMs grow in size and performance and are increasingly deployed in high-stakes applications, the need to understand and explain their behaviour becomes more crucial. Post-hoc explainability methods such as LIME \citep{ribeiro} are one way of attempting to do this. Although these methods have been widely applied to LLMs \citep{madsen}, to the best of our knowledge no research has been conducted on the impact of model size on the quality of these kinds of explanations. Here we begin to fill this gap by investigating the impact of model size on the quality of LIME explanations. We apply two approaches to assess the quality of explanations, namely faithfulness \citep{chan} and plausibility \citep{deyoung}. While faithfulness aims to measure the extent to which an explanation reflects the true internal decision processes, plausibility assesses the quality of the explanations based on their agreement with human-generated explanations.

We find that, even though model performance increases with model size, the agreement between human-generated and LIME-generated explanations does not. This indicates some extent of misalignment between the explanations and the true internal decision processes. Our findings also imply possible flaws in removal-based faithfulness metrics based on the NLP task which points to more general limitations for highlight-based post-hoc explainability in NLP such as lack of expressiveness. This study serves as a first attempt to understand how post-hoc explainability is affected by model size. We hope that this research encourages others to further explore this area and to that end we provide an extensible code repository\footnote{\texttt{https://github.com/henningheyen/Scalability-Of-LLM-Posthoc-Explanations}} for others to build on.

\section{Methodology}

\paragraph{Models and Datasets}
We use fine-tuned DeBERTaV3 models from Huggingface of four different sizes, ranging from 22 to 304 million parameters\footnote{For architectural specifics see Table \ref{tab:models} in the appendix.}. Note that state-of-the-art models exhibit parameter counts in the order of billions. Due to computational constraints, we could not experiment with larger models. The models were fine-tuned on two standard natural language inference (NLI) datasets, matched MNLI \citep{mnli} and SNLI \citep{snli}. Instead of SNLI, we use e-SNLI \citep{esnli} which extends SNLI by human annotated highlights indicating the most important tokens with respect to the label. Additionally, we apply the models in a zero-shot classification (ZSC) setting using the CoS-e \citep{cose} dataset. CoS-e consists of commonsense questions with five candidate labels where the candidate labels differ for each question\footnote{The Huggingface API internally transforms zero-shot classification problems to an NLI problem.}. Similarly to e-SNLI, CoS-e contains human annotated highlights\footnote{For examples from MNLI, e-SNLI and CoS-e refer to Table \ref{tab:MNLIexamples}, \ref{tab:eSNLIexamples} and \ref{tab:CoSeexamples} in the appendix.}.

\begin{figure}[ht]
    \centering
    \includegraphics[width=1\linewidth]{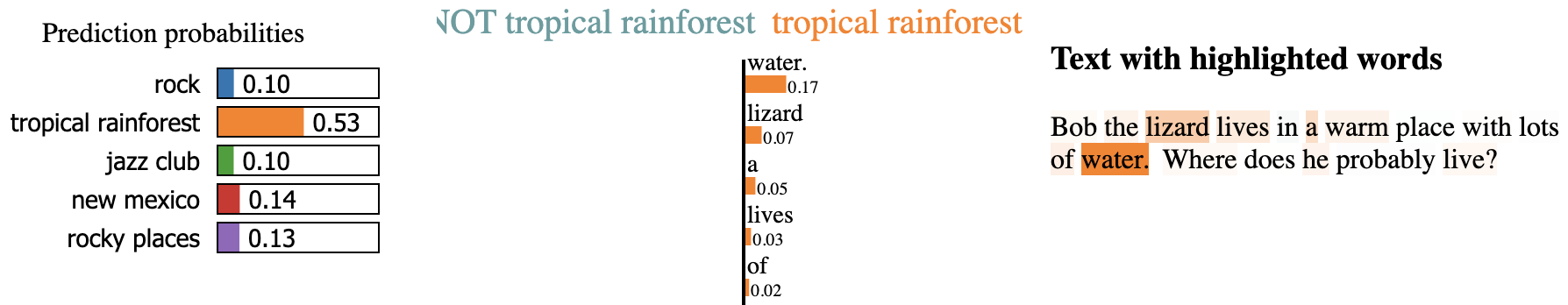}
    \caption{LIME example on a CoS-e instance using the xsmall DeBERTaV3 model. LIME maps every token to a real-valued importance score.}
    \label{fig:lime-example-cose}
\end{figure}

\paragraph{Explainability Method}
There exist different notions of explainability in NLP. One is in the form of free-text natural language explanations and another is in the form of highlight-based, typically post-hoc explanations. While the former approach commonly requires human evaluation the latter can be measured more objectively because post-hoc techniques in NLP are normally mappings from tokens to real-valued importance scores. Common techniques can be categorised as gradient-based, attention weight-based and perturbation-based. While gradient-based techniques are highly vulnerable to adversarial perturbation \citep{wang} several studies argue that explanations based on attention weights are unreliable. The study ``Attention is not Explanation`` \citep{jain}, for instance, identified different attention distributions yielding equivalent predictions. We believe perturbation-based methods avoid some of these pitfalls - we use one such method, LIME \citep{ribeiro}, in our experiments. LIME is a well-known and widely used explainability technique that is based on locally approximating the prediction with a simple interpretable model. The weights of the simple model serve as feature importance scores. Figure \ref{fig:lime-example-cose} illustrates a LIME explanation on a CoS-e instance. 

\paragraph{Explainability Metrics}
There is no single framework for evaluating post-hoc explanations due to a lack of consensus on what constitutes a high-quality explanation. In this work, we examine two different approaches, namely faithfulness and plausibility:

\textit{Faithfulness} as discussed in \cite{chan} aims to measure to what extent the explanation reflects the model’s internal decision process. Generally, faithfulness metrics rely on removing tokens from the input sequence based on the explanation and measuring the change in prediction. While several faithfulness metrics exist, they are ``not always consistent with each other and even lead to contradictory conclusions`` \citep{chan} - they compared six faithfulness metrics and concluded that \textbf{comprehensiveness} is the most diagnostic and the least complex. Based on this we use comprehensiveness for our experiments. Proposed by \cite{deyoung}, comprehensiveness suggests that an explanation is faithful if the prediction strongly deviates when the most important tokens (as identified by the explanation method) are removed from the input sequence\footnote{For a visualization of the comprehensiveness metric see Figure \ref{fig:faithfulness} in the appendix.}. More formally, 

       \begin{equation}
            \text{COMP}(\mathbf{x},c,k) = p(c \mid \mathbf{x}; \theta) - p(c \mid \mathbf{x} \backslash \mathbf{x}_{k}; \theta),
        \end{equation}
        \label{equ:COMP1}
        
where $p(c\mid \mathbf{x}; \theta)$ denotes the model's prediction for class $c$ on the entire input sequence and $p(c \mid\mathbf{x} \backslash \mathbf{x}_{k}; \theta)$ denotes the model's prediction when the top $k$ most important tokens $\mathbf{x}_k$ are removed from the input string. Practically, we obtain the most important tokens by taking the top $t$ percent tokens from the list of tokens with associated real-valued importance scores generated by a post-hoc explainability method such as LIME. We denote this explanation as $\mathbf{x}_t$. To enhance the metric’s reliability, the original paper proposes aggregated comprehensiveness which averages the comprehensiveness over different lengths of explanations.  In this work, we use bins $t \in T = \{10\%, 30\%, 50\%\}$ to vary the length of explanations. The aggregated comprehensiveness can be defined as, 

\begin{equation}
        \text{COMP}_{\text{agg}}(\mathbf{x}, c) = \sum_{t \in T} \text{COMP}(\mathbf{x}, c, t) = \sum_{t \in T} \left( p(c \mid \mathbf{x}; \theta) - p(c \mid \mathbf{x} \setminus \mathbf{x}_{t}; \theta) \right).
        \end{equation}
        \label{equ:aggCOMP1}
        
\textit{Plausibility}, as compared to faithfulness, defines the quality of an explanation by the intersection between the highlights generated by a post-hoc explainability technique and some human-annotated highlights. In other words, plausibility measures the ``agreement between extracted and human rationale`` \citep{deyoung}. Plausibility fundamentally differs from faithfulness in that plausible explanations do not reveal whether the model actually relied on the explanation. Assessing plausibility typically requires human evaluation \citep{strout}. However, more recently some existing datasets have been extended by human-annotated highlights \citep{deyoung} which allows for more quantitative evaluation of plausibility. In this paper, we use two datasets with human-annotated highlights, namely CoS-e \citep{cose} and e-SNLI \citep{esnli}. Similarly to \cite{deyoung} we measure plausibility by the \textbf{intersection over union (IOU)}. As proposed by \cite{deyoung}, we take the number of most important tokens according to the average explanation length provided by humans for each dataset\footnote{Mean explanation-input-ratio e-SNLI: $0.19$ ($\pm0.193$),  CoS-e: $0.26$ ($\pm0.137$).}. Suppose $\mathbf{x}_1$ is the set of tokens from the human explanation and $\mathbf{x}_2$ is the set of generated tokens. Then IOU can be formalised by

        \begin{equation}
        \label{equ:IOU1}
            \text{IOU}(\mathbf{x}_1, \mathbf{x}_2) = \frac{ \left| \mathbf{x}_1 \cap \mathbf{x}_2 \right| }{ \left| \mathbf{x}_1 \cup \mathbf{x}_2 \right| }.
        \end{equation}

\section{Experiments and Results}

\begin{figure}[ht]
    \centering
    \includegraphics[width=1\linewidth]{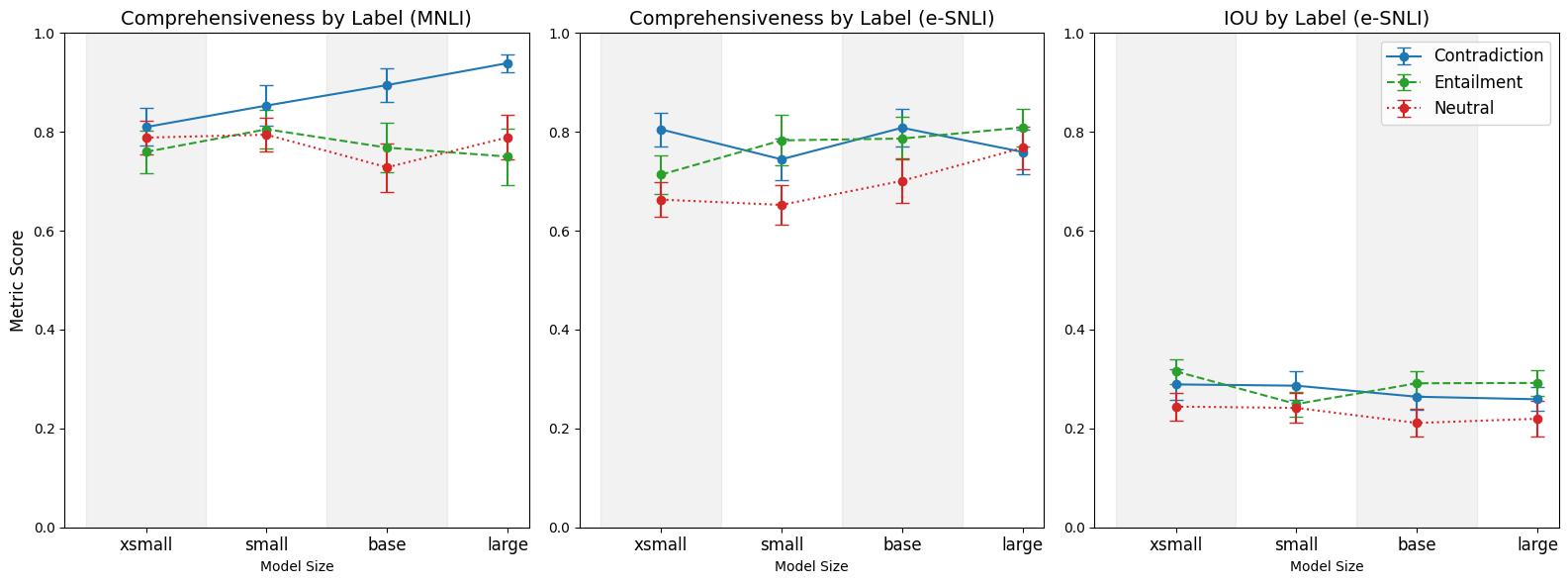}
    \caption{Mean comprehensiveness scores by labels for MNLI (left), e-SNLI (middle) and mean IOU scores by labels for e-SNLI (right) with mean standard errors on 100 test samples across all model sizes. Note how neutral sentence pairs achieve generally lower comprehensiveness scores than contradictive sentence pairs and how IOU scores are almost constant as the model size increases regardless of the label. Exact numbers are displayed in Table \ref{tab:faithfulness_and_plausibility_by_label} in the appendix.}
    \label{fig:all_by_label}
\end{figure}

The results for Experiment 1 and 2 are summarised in Table \ref{tab:combined_performance1}. Experiment 3 results are visualised in Figure \ref{fig:all_by_label}.

\textbf{Experiment 1} First, the four DeBERTaV3 models were evaluated in terms of performance on the validation sets of all three datasets (MNLI, e-SNLI, CoS-e). We report on model accuracy and 95\% confidence intervals. We find that, as expected, performance improves monotonically with increasing model size for all three datasets. We can conclude that the models' capabilities are different enough to reason about the effect of model size on the LIME explanations.

\textbf{Experiment 2} We then computed LIME explanations for each model on a subset of 100 test samples from each dataset. We had to use a subset due to the computational intensity of LIME. The explanations were always calculated with respect to the predicted class, not necessarily the correct class. For each explanation, the aggregated comprehensiveness and IOU scores were computed according to equation \ref{equ:aggCOMP1} and \ref{equ:IOU1} respectively. Note that IOU scores were only feasible for e-SNLI and CoS-e instances since MNLI does not provide human-annotated highlights. For each model, we report on the mean comprehensiveness and mean IOU scores across all 100 explanations. We observe an overall increase in comprehensiveness with the highest scores on the largest model for all three datasets suggesting that faithfulness of LIME increases with model size. IOU, on the other hand, stays almost constant across all model sizes for both datasets indicating that the plausibility of the LIME explanations is uncorrelated with model size. 

\newpage

\begin{table}[ht]
\centering
\footnotesize
\begin{tabular}{llllll}
\toprule
\textbf{Dataset} & \textbf{Model Size} & \textbf{Comprehensiveness} & \textbf{IOU} & \textbf{Accuracy} & \textbf{95\% C.I.} \\
\midrule
\multirow{4}{*}{MNLI}    
& xsmall & 0.785 ($\pm$ 0.022) & -- & 0.878 & (0.871, 0.885) \\
& small  & 0.817 ($\pm$ 0.022) & -- & 0.878 & (0.872, 0.884) \\
& base   & 0.796 ($\pm$ 0.027) & -- & 0.900 & (0.894, 0.906) \\
& large  & \textbf{0.823} ($\pm$ 0.027) & -- & \textbf{0.902} & (0.896, 0.908) \\
\midrule
\multirow{4}{*}{e-SNLI}  
& xsmall & 0.726 ($\pm$ 0.022) & \textbf{0.282} ($\pm$ 0.017) & 0.920 & (0.915, 0.925) \\
& small  & 0.724 ($\pm$ 0.026) & 0.259 ($\pm$ 0.016)          & 0.922 & (0.917, 0.927) \\
& base   & 0.764 ($\pm$ 0.025) & 0.254 ($\pm$ 0.016)          & 0.931 & (0.926, 0.936) \\
& large  & \textbf{0.778} ($\pm$ 0.025) & 0.256 ($\pm$ 0.017) & \textbf{0.932} & (0.927, 0.937) \\
\midrule
\multirow{4}{*}{CoS-e}   
& xsmall & 0.304 ($\pm$ 0.018) & 0.233 ($\pm$ 0.013) & 0.331 & (0.305, 0.355) \\
& small  & 0.316 ($\pm$ 0.019) & 0.231 ($\pm$ 0.014) & 0.336 & (0.306, 0.362) \\
& base   & 0.356 ($\pm$ 0.020)  & \textbf{0.235} ($\pm$ 0.012) & 0.359 & (0.330, 0.383) \\
& large  & \textbf{0.391} ($\pm$ 0.022) & 0.230 ($\pm$ 0.012) & \textbf{0.378} & (0.349, 0.406) \\
\bottomrule
\end{tabular}
\caption{Mean comprehensiveness and IOU scores with mean standard errors on 100 test samples for each dataset across all model sizes and accuracy scores on full validation sets with 95\% confidence intervals. IOU could not be computed on MNLI as this dataset does not provide human annotated highlights as ground truth explanations. }
\label{tab:combined_performance1}
\end{table}

\textbf{Experiment 3} Lastly, we investigated both metrics with respect to the labels (\textit{entailment, neutral, contradiction}) for MNLI and e-SNLI\footnote{Exact numbers are shown in Table \ref{tab:faithfulness_and_plausibility_by_label} in the appendix, the labels are balanced, see Table \ref{tab:label_counts}.}. The goal was to see how the metrics change with model size when we condition on the label. We observe that comprehensiveness improves for contradictory sentence pairs with larger model sizes in MNLI, while no consistent pattern emerges in e-SNLI. Generally, we find that neutral sentence pairs achieved lower comprehensiveness scores than contradiction pairs. IOU stays almost constant across different model sizes regardless of the label.  

\section{Discussion}

Overall, we find that for all three datasets, the largest model achieved the highest comprehensiveness score suggesting that with LIME larger models yield more faithful explanations. However, the IOU score stays constant across different model sizes suggesting that the plausibility of the explanations is uncorrelated with model size and performance. Interestingly, this would imply that the agreement with human-annotated highlights does
not improve with model performance which indicates an inherent misalignment between the
generated explanations and the true internal decision process. This finding seems contradictory to our previous result that with LIME larger models yield more faithful explanations. Splitting the metrics by labels could reveal potential flaws with the comprehensiveness metric in the NLI setting as we found significantly lower scores for neutral sentence pairs. The problem with comprehensiveness in an NLI setting could be that removing highlighted tokens from a neutral pair might very well result in another neutral prediction which limits the applicability of comprehensiveness in this case. More generally, this shows how post-hoc explanations in NLP lack expressiveness. Highlighting may not be sufficient to fully explain LLMs. While highlighting can provide some intuition about which tokens LLMs pay attention to, some higher-level reasoning concepts such as token dependencies or logical relations can hardly be expressed by highlighted tokens. This observation limits our finding that LIME explanations are more faithful for larger models. We conclude that the applicability of comprehensiveness is task-dependent and more coherent and expressive explainability metrics and techniques are needed.  

\section{Future Work}

The exclusive focus on LIME cannot capture the full range of challenges and opportunities that other techniques might present for explainability. Future research could repeat our experiments using other perturbation-based post-hoc techniques such as Anchors \citep{ribeiroanchors} or SHAP \citep{Lundberg} to validate our observations. Additionally, other tasks such as sentiment analysis, text summarisation or language modelling could be explored. Furthermore, none of our models have gone through Reinforcement Learning from Human Feedback (RLHF) \citep{rlhf}. We believe that the results on plausibility might significantly change with RLHF. While this work explored explainability primarily quantitatively, future investigations may benefit from qualitative analysis to deepen understanding of why discrepancies between model explanations and human understanding occur. Note that our large models had 304 million parameters. The current state-of-the-art models, however, are much bigger, consisting of billions of parameters. While computational feasibility with perturbation-based techniques is one concern, the effect on the explainability of very large models is still unexplored and should be the subject of future investigation. More broadly, this work aims to emphasize the urgent need for an LLM explainability framework that is human-interpretable, objective, scalable and expressive. A promising direction from a recent study showed empirically that the faithfulness of post hoc explanations could improve when the models are optimized for robustness against adversarial attacks \citep{li}. The method incorporates the explanations into the model training process.

\section{Conclusion}

Our work investigated LIME explanations on fine-tuned DeBERTaV3 models of different sizes in an NLI and ZSC setting. We applied two approaches to capture the quality of explanations, namely faithfulness and plausibility. We measured faithfulness with comprehensiveness and plausibility with IOU. We identified limitations of the comprehensiveness metric in the NLI setting which suggests that removal-based faithfulness metrics are task-dependent and that their general applicability is questionable. Given that performance increased with model size raises questions on why agreement with human annotations does not increase. We suggest that there is some extent of misalignment between the model’s internal decision process and its LIME explanation. This work aims to serve as an initial step towards understanding the effect of model size on LLM post-hoc explainability. We believe there is an urgent need for further investigations on LLM explainability and a more coherent explainability framework for LLMs. If we fail to faithfully explain the decisions of increasingly large language models we risk that those models pursue unexpected objectives rather than agreeing with human values and intentions.

\newpage

\bibliography{iclr2024_conference}

\begin{thebibliography}{17}
\providecommand{\natexlab}[1]{#1}
\providecommand{\url}[1]{\texttt{#1}}
\expandafter\ifx\csname urlstyle\endcsname\relax
  \providecommand{\doi}[1]{doi: #1}\else
  \providecommand{\doi}{doi: \begingroup \urlstyle{rm}\Url}\fi

\bibitem[Bowman et~al.(2015)Bowman, Angeli, Potts, and Manning]{snli}
Samuel~R. Bowman, Gabor Angeli, Christopher Potts, and Christopher~D. Manning.
\newblock A large annotated corpus for learning natural language inference.
\newblock In \emph{Proceedings of the 2015 Conference on Empirical Methods in Natural Language Processing}, pp.\  632--642, Lisbon, Portugal, September 2015. Association for Computational Linguistics.
\newblock \doi{10.18653/v1/D15-1075}.
\newblock URL \url{https://aclanthology.org/D15-1075}.

\bibitem[Brown et~al.(2020)Brown, Mann, Ryder, Subbiah, Kaplan, Dhariwal, Neelakantan, Shyam, Sastry, Askell, et~al.]{brownGPT3}
Tom Brown, Benjamin Mann, Nick Ryder, Melanie Subbiah, Jared~D Kaplan, Prafulla Dhariwal, Arvind Neelakantan, Pranav Shyam, Girish Sastry, Amanda Askell, et~al.
\newblock Language models are few-shot learners.
\newblock \emph{Advances in neural information processing systems}, 33:\penalty0 1877--1901, 2020.

\bibitem[Camburu et~al.(2018)Camburu, Rockt{\"a}schel, Lukasiewicz, and Blunsom]{esnli}
Oana-Maria Camburu, Tim Rockt{\"a}schel, Thomas Lukasiewicz, and Phil Blunsom.
\newblock e-snli: Natural language inference with natural language explanations.
\newblock \emph{Advances in Neural Information Processing Systems}, 31, 2018.

\bibitem[Chan et~al.(2022)Chan, Kong, and Guanqing]{chan}
Chun~Sik Chan, Huanqi Kong, and Liang Guanqing.
\newblock A comparative study of faithfulness metrics for model interpretability methods.
\newblock In \emph{Proceedings of the 60th Annual Meeting of the Association for Computational Linguistics (Volume 1: Long Papers)}, pp.\  5029--5038, Dublin, Ireland, May 2022. Association for Computational Linguistics.
\newblock \doi{10.18653/v1/2022.acl-long.345}.
\newblock URL \url{https://aclanthology.org/2022.acl-long.345}.

\bibitem[DeYoung et~al.(2020)DeYoung, Jain, Rajani, Lehman, Xiong, Socher, and Wallace]{deyoung}
Jay DeYoung, Sarthak Jain, Nazneen~Fatema Rajani, Eric Lehman, Caiming Xiong, Richard Socher, and Byron~C. Wallace.
\newblock {ERASER}: {A} benchmark to evaluate rationalized {NLP} models.
\newblock In \emph{Proceedings of the 58th Annual Meeting of the Association for Computational Linguistics}, pp.\  4443--4458, Online, July 2020. Association for Computational Linguistics.
\newblock \doi{10.18653/v1/2020.acl-main.408}.
\newblock URL \url{https://aclanthology.org/2020.acl-main.408}.

\bibitem[Jain \& Wallace(2019)Jain and Wallace]{jain}
Sarthak Jain and Byron~C. Wallace.
\newblock {A}ttention is not {E}xplanation.
\newblock In \emph{Proceedings of the 2019 Conference of the North {A}merican Chapter of the Association for Computational Linguistics: Human Language Technologies, Volume 1 (Long and Short Papers)}, pp.\  3543--3556, Minneapolis, Minnesota, June 2019. Association for Computational Linguistics.
\newblock \doi{10.18653/v1/N19-1357}.
\newblock URL \url{https://aclanthology.org/N19-1357}.

\bibitem[Kaplan et~al.(2020)Kaplan, McCandlish, Henighan, Brown, Chess, Child, Gray, Radford, Wu, and Amodei]{kaplan}
Jared Kaplan, Sam McCandlish, Tom Henighan, Tom~B Brown, Benjamin Chess, Rewon Child, Scott Gray, Alec Radford, Jeffrey Wu, and Dario Amodei.
\newblock Scaling laws for neural language models.
\newblock \emph{arXiv preprint arXiv:2001.08361}, 2020.

\bibitem[Li et~al.(2023)Li, Hu, Chen, and He]{li}
Dongfang Li, Baotian Hu, Qingcai Chen, and Shan He.
\newblock Towards faithful explanations for text classification with robustness improvement and explanation guided training.
\newblock In Anaelia Ovalle, Kai-Wei Chang, Ninareh Mehrabi, Yada Pruksachatkun, Aram Galystan, Jwala Dhamala, Apurv Verma, Trista Cao, Anoop Kumar, and Rahul Gupta (eds.), \emph{Proceedings of the 3rd Workshop on Trustworthy Natural Language Processing (TrustNLP 2023)}, pp.\  1--14, Toronto, Canada, July 2023. Association for Computational Linguistics.
\newblock \doi{10.18653/v1/2023.trustnlp-1.1}.
\newblock URL \url{https://aclanthology.org/2023.trustnlp-1.1}.

\bibitem[Lundberg \& Lee(2017)Lundberg and Lee]{Lundberg}
Scott~M. Lundberg and Su-In Lee.
\newblock A unified approach to interpreting model predictions.
\newblock In \emph{Proceedings of the 31st International Conference on Neural Information Processing Systems}, NIPS'17, pp.\  4768--4777, Red Hook, NY, USA, 2017. Curran Associates Inc.
\newblock ISBN 9781510860964.

\bibitem[Madsen et~al.(2022)Madsen, Reddy, and Chandar]{madsen}
Andreas Madsen, Siva Reddy, and Sarath Chandar.
\newblock Post-hoc interpretability for neural nlp: A survey.
\newblock \emph{ACM Computing Surveys}, 55\penalty0 (8):\penalty0 1--42, 2022.

\bibitem[Ouyang et~al.(2022)Ouyang, Wu, Jiang, Almeida, Wainwright, Mishkin, Zhang, Agarwal, Slama, Ray, et~al.]{rlhf}
Long Ouyang, Jeffrey Wu, Xu~Jiang, Diogo Almeida, Carroll Wainwright, Pamela Mishkin, Chong Zhang, Sandhini Agarwal, Katarina Slama, Alex Ray, et~al.
\newblock Training language models to follow instructions with human feedback.
\newblock \emph{Advances in Neural Information Processing Systems}, 35:\penalty0 27730--27744, 2022.

\bibitem[Rajani et~al.(2019)Rajani, McCann, Xiong, and Socher]{cose}
Nazneen~Fatema Rajani, Bryan McCann, Caiming Xiong, and Richard Socher.
\newblock Explain yourself! leveraging language models for commonsense reasoning.
\newblock In \emph{Proceedings of the 57th Annual Meeting of the Association for Computational Linguistics}, pp.\  4932--4942, Florence, Italy, July 2019. Association for Computational Linguistics.
\newblock \doi{10.18653/v1/P19-1487}.
\newblock URL \url{https://aclanthology.org/P19-1487}.

\bibitem[Ribeiro et~al.(2016)Ribeiro, Singh, and Guestrin]{ribeiro}
Marco~Tulio Ribeiro, Sameer Singh, and Carlos Guestrin.
\newblock " why should i trust you?" explaining the predictions of any classifier.
\newblock In \emph{Proceedings of the 22nd ACM SIGKDD international conference on knowledge discovery and data mining}, pp.\  1135--1144, 2016.

\bibitem[Ribeiro et~al.(2018)Ribeiro, Singh, and Guestrin]{ribeiroanchors}
Marco~Tulio Ribeiro, Sameer Singh, and Carlos Guestrin.
\newblock Anchors: High-precision model-agnostic explanations.
\newblock In \emph{Proceedings of the AAAI conference on artificial intelligence}, volume~32, 2018.

\bibitem[Strout et~al.(2019)Strout, Zhang, and Mooney]{strout}
Julia Strout, Ye~Zhang, and Raymond Mooney.
\newblock Do human rationales improve machine explanations?
\newblock In \emph{Proceedings of the 2019 ACL Workshop BlackboxNLP: Analyzing and Interpreting Neural Networks for NLP}, pp.\  56--62, Florence, Italy, August 2019. Association for Computational Linguistics.
\newblock \doi{10.18653/v1/W19-4807}.
\newblock URL \url{https://aclanthology.org/W19-4807}.

\bibitem[Wang et~al.(2020)Wang, Tuyls, Wallace, and Singh]{wang}
Junlin Wang, Jens Tuyls, Eric Wallace, and Sameer Singh.
\newblock Gradient-based analysis of {NLP} models is manipulable.
\newblock In \emph{Findings of the Association for Computational Linguistics: EMNLP 2020}, pp.\  247--258, Online, November 2020. Association for Computational Linguistics.
\newblock \doi{10.18653/v1/2020.findings-emnlp.24}.
\newblock URL \url{https://aclanthology.org/2020.findings-emnlp.24}.

\bibitem[Williams et~al.(2018)Williams, Nangia, and Bowman]{mnli}
Adina Williams, Nikita Nangia, and Samuel Bowman.
\newblock A broad-coverage challenge corpus for sentence understanding through inference.
\newblock In \emph{Proceedings of the 2018 Conference of the North {A}merican Chapter of the Association for Computational Linguistics: Human Language Technologies, Volume 1 (Long Papers)}, pp.\  1112--1122, New Orleans, Louisiana, June 2018. Association for Computational Linguistics.
\newblock \doi{10.18653/v1/N18-1101}.
\newblock URL \url{https://aclanthology.org/N18-1101}.

\end{thebibliography}
\bibliographystyle{iclr2024_conference}

\newpage
\appendix
\section{Appendix}
\subsection{Additional Tables}

\begin{table}[ht]
\centering
\footnotesize
\begin{tabular}{@{}l l l l l@{}}
\toprule
\textbf{Dataset} & \textbf{Model Size} & \textbf{Label} & \textbf{Comprehensiveness} & \textbf{IOU} \\
\midrule
\multirow{12}{*}{MNLI}   
& \multirow{3}{*}{xsmall} & contradiction & 0.810 ($\pm$ 0.038) & - \\
&                         & entailment    & 0.759 ($\pm$ 0.042) & - \\
&                         & neutral       & 0.788 ($\pm$ 0.034) & - \\
\addlinespace
& \multirow{3}{*}{small}  & contradiction & 0.853 ($\pm$ 0.041) & - \\
&                         & entailment    & 0.805 ($\pm$ 0.039) & - \\
&                         & neutral       & 0.794 ($\pm$ 0.035) & - \\
\addlinespace
& \multirow{3}{*}{base}   & contradiction & 0.895 ($\pm$ 0.034) & - \\
&                         & entailment    & 0.768 ($\pm$ 0.050) & - \\
&                         & neutral       & 0.728 ($\pm$ 0.049) & - \\
\addlinespace
& \multirow{3}{*}{large}  & contradiction & 0.939 ($\pm$ 0.018) & - \\
&                         & entailment    & 0.750 ($\pm$ 0.057) & - \\
&                         & neutral       & 0.789 ($\pm$ 0.046) & - \\
\midrule
\multirow{12}{*}{e-SNLI}  
& \multirow{3}{*}{xsmall} & contradiction & 0.805 ($\pm$ 0.034) & 0.289 ($\pm$ 0.031) \\
&                         & entailment    & 0.713 ($\pm$ 0.039) & 0.315 ($\pm$ 0.025) \\
&                         & neutral       & 0.663 ($\pm$ 0.035) & 0.244 ($\pm$ 0.028) \\
\addlinespace
& \multirow{3}{*}{small}  & contradiction & 0.744 ($\pm$ 0.042) & 0.286 ($\pm$ 0.029) \\
&                         & entailment    & 0.783 ($\pm$ 0.051) & 0.249 ($\pm$ 0.025) \\
&                         & neutral       & 0.652 ($\pm$ 0.040) & 0.242 ($\pm$ 0.030) \\
\addlinespace
& \multirow{3}{*}{base}   & contradiction & 0.808 ($\pm$ 0.038) & 0.264 ($\pm$ 0.027) \\
&                         & entailment    & 0.786 ($\pm$ 0.043) & 0.291 ($\pm$ 0.025) \\
&                         & neutral       & 0.701 ($\pm$ 0.045) & 0.211 ($\pm$ 0.028) \\
\addlinespace
& \multirow{3}{*}{large}  & contradiction & 0.759 ($\pm$ 0.046) & 0.259 ($\pm$ 0.024) \\
&                         & entailment    & 0.809 ($\pm$ 0.038) & 0.292 ($\pm$ 0.026) \\
&                         & neutral       & 0.768 ($\pm$ 0.042) & 0.220 ($\pm$ 0.036) \\
\bottomrule
\end{tabular}
\caption{Mean comprehensiveness and IOU scores with mean standard errors on 100 test samples split by label for both NLI datasets across all model sizes. IOU could not be computed on MNLI as this dataset does not provide human-annotated highlights as ground truth explanations.}
\label{tab:faithfulness_and_plausibility_by_label}
\end{table}

\begin{table}[ht]
\centering
\begin{tabular}{lcccc}
\toprule
 & \multicolumn{1}{c}{\textbf{MNLI}} & \multicolumn{1}{c}{\textbf{e-SNLI}} & \multicolumn{1}{c}{\textbf{CoS-e}} \\
\midrule
xsmall & 2min 3s & 1min 8s & 34min 35s \\
small  & 2min 40s & 1min 40s & 44min 28s \\
base   & 5min 20s & 3min 35s & 1h 27min 7s \\
large  & 15min 38s & 12min 45s & 4h 35min 50s \\
\bottomrule
\end{tabular}
\caption{Compute time for all LIME explanations of 100 test instances from each dataset across all model sizes on Nvidia's T4 GPU, 51GB RAM.}
\label{tab:compute_time}
\end{table}

\begin{table}[ht]
\centering
\footnotesize
\begin{tabular}{lcccc}
\toprule
 & \textbf{Parameters} & \textbf{Layers} & \textbf{Hidden} & \textbf{Attention} \\
 & \textbf{(in millions)} & & \textbf{Size} & \textbf{Heads} \\
\midrule
large & 304 & 24 & 1024 & 12 \\
base & 86 & 12 & 768 & 12 \\
small & 44 & 6 & 768 & 12 \\
xsmall & 22 & 12 & 384 & 6\\
\bottomrule
\end{tabular}
\caption{Architecture comparison for DeBERTaV3 models.}
\label{tab:models}
\end{table}

\begin{table}[ht]
\centering
\footnotesize
\begin{tabular}{P{6cm}P{5cm}P{2cm}} 
\toprule
\textbf{Premise} & \textbf{Hypothesis} & \textbf{Label} \\
\midrule
Look, there's a legend here. & See, there is a well-known hero here. &  Entailment \\
\addlinespace
Yeah, I know, and I did that all through college and it worked too. & I did that all through college but it never worked. & Contradiction \\
\addlinespace
Boats in daily use lie within feet of the fashionable bars and restaurants. & Bars and restaurants are interesting places.  & Neutral \\
\bottomrule
\end{tabular}
\caption{Natural language inference examples from the MNLI dataset.}
\label{tab:MNLIexamples}
\end{table}

\begin{table}[ht]
\centering
\footnotesize
\begin{tabular}{P{6cm}P{5cm}P{2cm}}
\toprule
\textbf{Premise} & \textbf{Hypothesis} & \textbf{Label} \\
\midrule
An adult dressed in black \colorbox{yellow}{holds a stick}. & An adult is walking away, \colorbox{yellow}{empty-handed}. & Contradiction \\
\addlinespace
A child in a yellow plastic safety swing is laughing as a dark-haired woman in pink and coral pants stands behind her. & A young \colorbox{yellow}{mother} is playing with her \colorbox{yellow}{daughter} in a swing. & Neutral \\
\addlinespace
A \colorbox{yellow}{man} in an orange vest \colorbox{yellow}{leans over a pickup truck}. & A man is \colorbox{yellow}{touching} a truck. & Entailment \\
\bottomrule
\end{tabular}
\caption{Natural language inference examples from the e-SNLI dataset. Highlighted tokens indicate human-annotated explanations.}
\label{tab:eSNLIexamples}
\end{table}

\begin{table}[ht]
\centering
\footnotesize
\begin{tabular}{P{6cm}P{5cm}P{2cm}}
\toprule
\textbf{Question} & \textbf{Candidate Labels} & \textbf{Label} \\
\midrule
He was a \colorbox{yellow}{sloppy eater}, so where did he leave a mess?
 & sailboat, desk, closet, table, apartment & table \\
\addlinespace 
Where can someone get a \colorbox{yellow}{new saw}? & hardware store, toolbox, logging camp, tool kit, auger & hardware store \\
\addlinespace 
Many homes \colorbox{yellow}{in this country} are built around a courtyard. Where is it? & hospital, park, spain, office complex, office & spain \\
\bottomrule
\end{tabular}
\caption{Zero shot classification examples from the CoS-e dataset. Highlighted tokens indicate human-annotated explanations.}
\label{tab:CoSeexamples}
\end{table}

\begin{table}[ht]
\centering
\begin{tabular}{lccc}
\toprule
\textbf{Dataset} & \textbf{Contradiction} & \textbf{Entailment} & \textbf{Neutral} \\
\midrule
MNLI & 32 & 36 & 32 \\
e-SNLI & 33 & 32 & 35 \\
\bottomrule
\end{tabular}
\caption{Number of observations by label for MNLI and e-SNLI for 100 explained test samples.}
\label{tab:label_counts}
\end{table} 

\clearpage
\subsection{Additional Figures}

\begin{figure}[ht]
    \centering
    \includegraphics[width=0.95\linewidth]{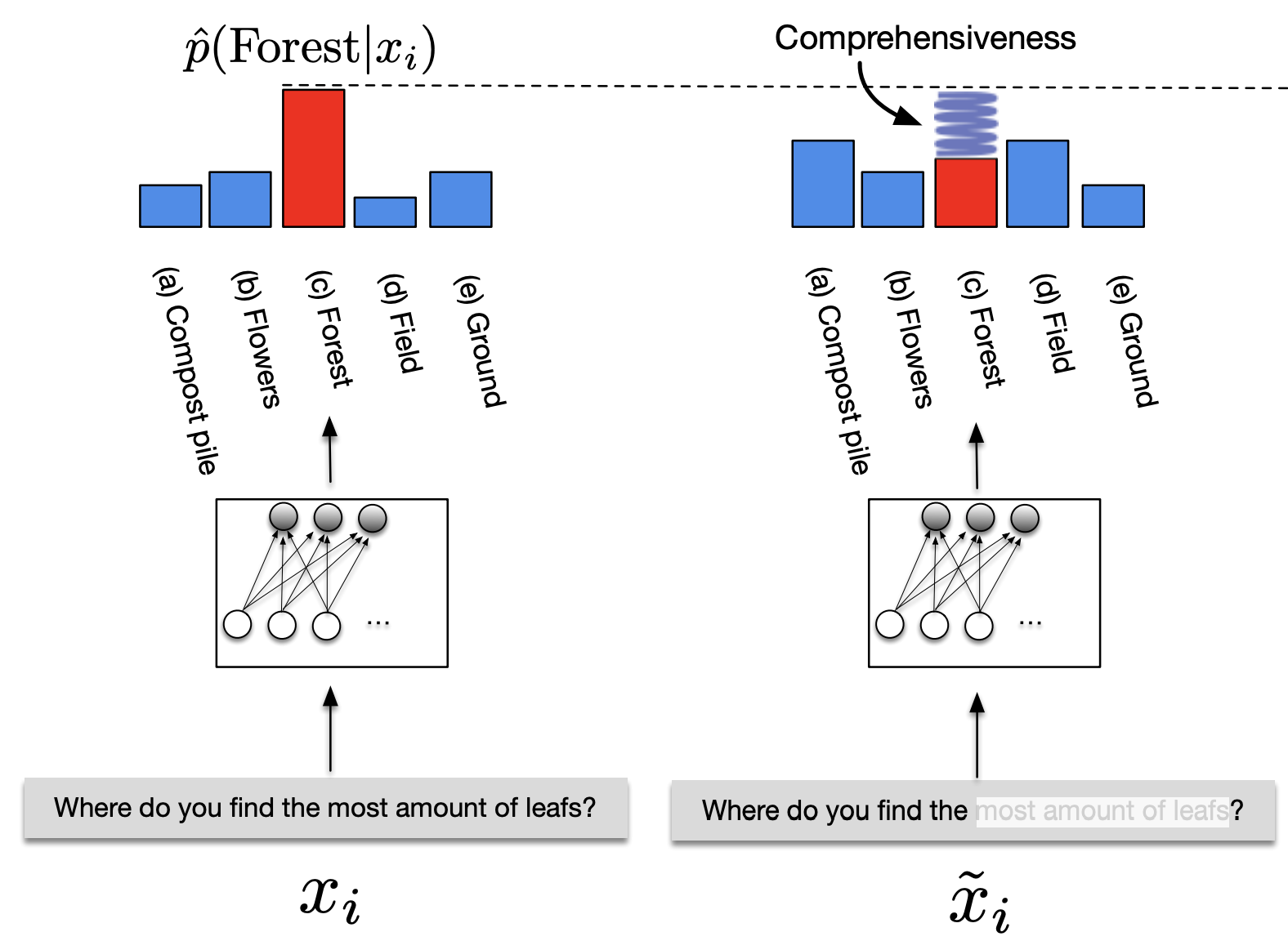}
    \caption{Visualistion of comprehensiveness metric on CoS-e instance from \citep{deyoung}. Comprehensiveness suggests that an explanation is faithful if the prediction strongly deviates when the most important tokens (as identified by the explanation method) are removed from the input sequence.}
    \label{fig:faithfulness}
\end{figure}

\begin{figure}[ht]
    \centering
    \includegraphics[width=0.95\linewidth]{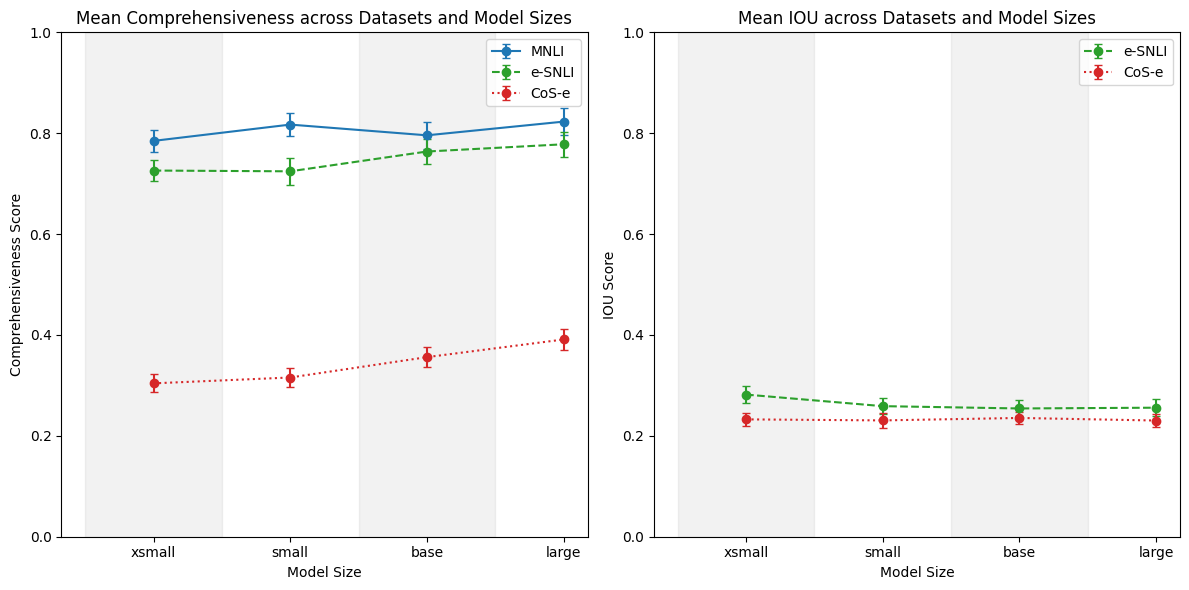}
    \caption{Mean comprehensiveness (left) and IOU (right) scores with mean standard errors on 100 test samples for each dataset across all model sizes. IOU could not be computed on MNLI as this dataset does not provide human-annotated highlights as ground truth explanations.}
    \label{fig:global_comprehensiveness_and_iou}
\end{figure}

\end{document}